%% file: main.tex
\pdfoutput=1

\documentclass[11pt]{article}

\usepackage[]{EMNLP2023}

\usepackage{times}
\usepackage{latexsym}

\usepackage[T1]{fontenc}

\usepackage[utf8]{inputenc}

\usepackage{microtype}

\usepackage{inconsolata}

\usepackage{amsmath,amsfonts,amssymb,amsthm,mathtools}
\input{math_commands.tex}

\usepackage{times}
\usepackage{latexsym}
\usepackage{microtype}
\usepackage{etoolbox}
\usepackage{times}
\usepackage{latexsym}
\usepackage{comment}
\usepackage{float}
\usepackage{capt-of}
\usepackage{caption}
\usepackage{siunitx}
\usepackage{fontawesome}
\usepackage{booktabs}
\usepackage{graphicx}
\usepackage{xcolor}
\usepackage{tikz-dependency}
\usepackage{booktabs, tabularx}
\usepackage{subfigure}
\usepackage{fancyvrb}

\usepackage{color,soul}
\usepackage{linguex}
\usepackage{array}
\usepackage{bm}
\usepackage{arydshln}
\usepackage{tikz}
\usepackage{multirow}
\usepackage{color}
\usepackage{colortbl}
\usepackage{longtable}
\usepackage[inline]{enumitem}
\usepackage{lipsum}
\usepackage{xurl}
\makeatletter
\def\adl@drawiv#1#2#3{%
        \hskip.5\tabcolsep
        \xleaders#3{#2.5\@tempdimb #1{1}#2.5\@tempdimb}%
                #2\z@ plus1fil minus1fil\relax
        \hskip.5\tabcolsep}
\newcommand{\cdashlinelr}[1]{%
  \noalign{\vskip 1.3pt
           \global\let\@dashdrawstore\adl@draw
           \global\let\adl@draw\adl@drawiv}
  \cdashline{#1}[.4pt/2pt]
  \noalign{\global\let\adl@draw\@dashdrawstore
           \vskip 1.3pt}}

\def\blfootnote{\xdef\@thefnmark{}\@footnotetext}
\usepackage[ruled,vlined]{algorithm2e}

\newcommand{\ms}[1]{\textcolor{orange}{\bf\small }}
\newcommand{\hs}[1]{\textcolor{orange}{\bf\small }}
\newcommand{\jf}[1]{\textcolor{blue}{\bf\small }}

\definecolor{lightgreen}{RGB}{223,255,219}
\definecolor{lightred}{RGB}{255,219,219}
  
\input{macros}

\title{Automating Behavioral Testing in Machine Translation}

\author{Javier Ferrando$^{\diamondsuit}$\thanks{\, \, Work done during an internship at Apple.}\quad Matthias Sperber${^\dagger}$\quad  Hendra Setiawan${^\dagger}$\\ \textbf{Dominic Telaar${^\dagger}$\quad Saša Hasan${^\dagger}$} \\ 
$^{\diamondsuit}$Universitat Politècnica de Catalunya \\
${^\dagger}$Apple\\
\normalsize{\texttt{javier.ferrando.monsonis@upc.edu,sperber@apple.com}}
}

\begin{document}
\maketitle

\begin{abstract}
Behavioral testing in NLP allows fine-grained evaluation of systems by examining their linguistic capabilities through the analysis of input-output behavior. Unfortunately, existing work on behavioral testing in Machine Translation (MT) is currently restricted to largely handcrafted tests covering a limited range of capabilities and languages. To address this limitation, we propose to use Large Language Models (LLMs) to generate a diverse set of source sentences tailored to test the behavior of MT models in a range of situations. 
We can then verify whether the MT model exhibits the expected behavior through matching candidate sets that are also generated using LLMs.
Our approach aims to make behavioral testing of MT systems practical while requiring only minimal human effort.
In our experiments, we apply our proposed evaluation framework to assess multiple available MT systems, revealing that while in general pass-rates follow the trends observable from traditional accuracy-based metrics, our method was able to uncover several important differences and potential bugs that go unnoticed when relying only on accuracy.\footnote{Prompts and generated data are available at \url{https://github.com/apple/ml-behavioral-testing-for-mt}.}
\end{abstract}

\section{Introduction}
Automatic evaluation metrics such as BLEU \citep{papineni-etal-2002-bleu} and COMET \citep{rei-etal-2020-comet} are the primary means of measuring the translation quality of MT systems. Researchers and practitioners rely on them for comparing systems, detecting regressions, and making deployment decisions. This poses an important concern: such metrics typically aggregate the performance of systems across a set of sentences into single scores. Unfortunately, these metrics by design tend to overlook specific infrequent but important error cases, making it difficult to reliably detect such issues in practice.

\begin{table}[t]
\resizebox{0.49\textwidth}{!}{
\begin{tabular}{ll}
\toprule
\textbf{Property} &  \textbf{Translation Errors}    \\
\midrule
Integers &  7000000 $\to$ \textcolor{red}{70.000.000} \\
Decimals &  500.75 $\to$ 500\textcolor{red}{.}75 \\
Large Numbers & 1.366 billion $\to$ 1\textcolor{red}{.}366 Milliarden \\
Idioms &  ins and outs $\to$ \textcolor{red}{Ins und Outs} \\
Currencies  &  BRL $\to$ \textcolor{red}{RL} \\
Physical Units &  miles $\to$ \textcolor{red}{km} \\
\multirow{2}{*}{Web Terms}  & www.onlinegrocery.com $\to$\\
& www.online\textcolor{red}{e}grocery.com   \\
\multicolumn{1}{c}{...} & \multicolumn{1}{c}{...}\\
\bottomrule
\end{tabular}
}
\caption{Subset of linguistic properties tested with our proposed method, and examples (source $\rightarrow$ translation) of translation errors found in En$\rightarrow$De MT models.}
\label{tab:spec_table}
\vspace{-1.0em}
\end{table}

Behavioral testing, originally developed as a type of software testing ~\citep{behavioral_testing}, has been proposed as an approach that can alleviate such kinds of problems in natural language processing \citep{ribeiro-etal-2020-beyond}. Behavioral tests focus on assessing a system's fine-grained linguistic capabilities by validating input-output behavior in a controlled fashion.

\begin{figure*}[t]\begin{centering}\includegraphics[width=0.85\textwidth]{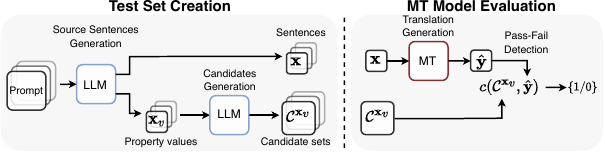}
	\caption{Pipeline of the proposed approach. Left: For each property type, a test set is created via a LLM, composed of source sentences $\mathbf{x}$ with property values $\mathbf{x}_v$ (\Cref{sec:source_sent_generation}). Subsequently, a candidate set of valid translations of each property value $\mathcal{C}^{\mathbf{x}_v}$ is generated (\Cref{sec:candidates_generation}). Right: During evaluation, the translation $\mathbf{\hat{y}}$ generated by an MT model is compared against the candidate sets, and a pass-fail decision is made (\Cref{sec:pass_fail_detector}).}
	\label{fig:project_pipeline}
	\end{centering}
\end{figure*}

\Cref{tab:spec_table} shows examples of typical issues of MT systems that could be covered by behavioral tests. We argue that the availability of a comprehensive behavioral test suite for MT would be of high practical value: It would allow understanding how exactly two MT models differ, or to block an MT system from being deployed if a passing threshold for a certain linguistic capability is not met.

However, there are currently two major limitations that arise when attempting to apply behavioral testing to MT. First of all, behavioral testing was originally designed for evaluating systems characterized by a relatively small output space. For instance, \citet{ribeiro-etal-2020-beyond} investigate sentiment classification, duplicate question detection, and machine comprehension. In contrast, the output space of MT systems grows exponentially as tokens are generated. Secondly, behavioral testing often requires rigid templates to create examples and their corresponding labels, which involves a costly human effort to develop and expand to additional use cases. Otherwise, the diversity of sentences in the resulting test suite is too limited.

Several recent works have partially addressed these limitations. For example, \citet{wang-etal-2021-easy,raunak-etal-2022-salted} propose MT-specific test sets which include the ability to handle large output space. \citet{yang-etal-2022-testaug} address the limitation of rigid templates. To the best of our knowledge, no prior work has addressed both limitations for MT.

In this study, we aim to bridge this gap by leveraging LLMs with in-context learning to automate the creation of behavioral tests in MT for the first time. Our main contributions  are as follows:
\begin{itemize}
    \item We use LLMs to automate the generation of a diverse set of source sentences for behavioral testing. Sentences are generated to exhibit the specific language property that is being tested.
    \item We verify whether an MT system's output contains an accurate translation of the language property that is being tested. To this end, we propose using LLMs to generate candidate sets of ground-truth translations of the property values in cases where exhaustive candidate sets are plausible. Otherwise, we generate contrastive candidates and evaluate via semantic similarity measures.
    \item We present an evaluation framework to robustly compute pass rates of MT models across various language properties, and show results for widely used open-source models on three language pairs.
\end{itemize}

\begin{figure*}[!t]
\input{prompt_src_no_format.tex}
\caption{General template of the prompt used for generating batches of source sentences.}
\label{fig:prompt_src}
\end{figure*}

\section{Behavioral Testing for MT}

Behavioral testing, as proposed by ~\citet{ribeiro-etal-2020-beyond}, uses input-output pairs tailored to evaluate a model capability to correctly handle certain language properties. 
The goal is to complement traditional aggregated accuracy scores, which, while useful by themselves, often fail to capture long-tail phenomena. In practice, manual inspection of system outputs is often crucial to make up for this shortcoming. Automated behavioral testing provides a more reliable and less cumbersome alternative that can reduce or eliminate the need for manual inspection, provided that a sufficient range of language properties is tested. Test results are presented in the form of a table of pass rates (one pass rate for each tested property) that is informative to decide on consequent steps, e.g.\ whether bugs must be addressed before deployment. Note that the creation of a \textit{sufficiently comprehensive} behavioral test suite depends crucially on whether its creation can be automated to a high degree, which is also our main design goal in this work.

We are particularly interested in a specific type of behavioral tests, \textit{minimal functionality tests} (MFTs) ~\cite{ribeiro-etal-2020-beyond}.\footnote{~\citet{ribeiro-etal-2020-beyond} also propose \textit{directional} and \textit{invariance} tests which check how model outputs change under certain input perturbance, but these appear less applicable to MT given the potentially large space of correct translations.}
In the context of MT, an MFT measures a model's ability to translate particular property values that appear naturally embedded in some given source sentences.

\Cref{fig:project_pipeline} illustrates our proposed framework. First, a source sentence ${\mathbf{x}=\{\vx_1\,\cdots,\vx_{|\mathbf{x}|}\}}$ that contains a tagged property value ${\mathbf{x}_v}{\subseteq}\mathbf{x}$ is generated (\Cref{sec:source_sent_generation}). For instance, if our test property is physical unit translation, we might have ${\mathbf{x}}{=}\textit{``I ran 3 miles.''}$ and ${\mathbf{x}_v}{=}\textit{miles}$. A main challenge comes from the fact that there is a potentially large space of correct translations. However, note that by design MFTs only need to check whether the property under test is translated correctly, while unrelated translation errors should be ignored. In many cases, this reduces the space of correct translations to a manageable size. We therefore propose to automatically generate a candidate set $\mathcal{C}^{\mathbf{x}_v}$ (either exhaustive or contrastive; see \Cref{sec:candidates_generation}) and then apply a pass-fail detector that uses either string matching or semantic similarity measures (\Cref{sec:pass_fail_detector}). In our example, we might generate an exhaustive candidate set $\mathcal{C}^{\mathbf{x}_v}{=}\{\textit{Meilen}, \textit{mi}\}$ for the case of translating into German. We now aim to evaluate an MT model ${f: \mathbf{x} \mapsto \mathbf{\hat{y}}}$. To do so, we compare $\mathbf{\hat{y}}$ against  $\mathcal{C}^{\mathbf{x}_v}$. A correct translation $\mathbf{\hat{y}}{=}\textit{``Ich lief 3 Meilen.''}$ would match the candidate set and therefore pass the test, while a typical incorrect translation $\mathbf{\hat{y}}{=}\textit{``Ich lief 3 km.''}$ does not match the candidate set and therefore fails the test.

Given this general overview of our method, we now turn to a more precise description of each proposed step in the following sections.

\section{Source Sentence Generation}
\label{sec:source_sent_generation}

To create source sentences for testing a certain language property, we pose several desiderata: Sentences should be \textit{diverse} (e.g.\ not rely only on a handful of templates), \textit{natural}, \textit{numerous} enough to yield statistical significance, and \textit{contain a property value} associated with our tested property. 

Note that existing approaches often struggle to generate diverse test sets due to the reliance on hand-crafted templates~\citep{wang-etal-2021-easy}. To overcome this shortcoming, we design a general template for prompting LLMs, in our case ChatGPT\footnote{\texttt{gpt-3.5-turbo} API accessed on May 2023.}, OpenAI's model built on InstructGPT \citep{ouyang2022training}. This allows us to generate diverse source language sentences that contain property values suitable for testing different capabilities (see prompt\footnote{We set temperature=0.9, presence\_penalty=2.} in \Cref{fig:prompt_src}). We instantiate the prompt once for every language property that we wish to include in our test suite.

To simplify the later verification step, we generate sentences that contain exactly one such property value $\mathbf{x}_v$.\footnote{For some types of properties, multiple property values may be more appropriate. This is left for future work.} We generate source sentences with brackets around the property value for easy parsing. A possible test sentence for the property of translating decimal numbers might look as follows:
\begin{equation}\label{ex:source_sentence}
{\text{The company received [}}\overbracket[0.5pt][7pt]{\text{4200.4}}^{\text{\tiny{property value}}}{\text{]\small{€}.}}
\end{equation}
Note that brackets are removed before passing the sentence to the MT model.

We apply basic filters to remove duplicated sentences, examples with more than one property value, or those composed of more than one sentence. We repeatedly feed the same prompt to the LLM, and stop the generation process when reaching 1,000 sentences after filtering. Our experiments (\Cref{sec:reliability}) indicate that ChatGPT is able to generate sentences of adequate quality and diversity.

\begin{table}[t]
    \begin{tabularx}{\linewidth}{lr}
        \toprule
\textbf{Candidates Examples}    \\
        \midrule
kilometers $\rightarrow$ kilómetros, km      \\
watts $\rightarrow$ vatios, W   \\
meters per second $\rightarrow$ metros por segundo, m/s\\
        \bottomrule
    \end{tabularx}
    \caption{Examples of En$\rightarrow$Es set of candidates generated by ChatGPT.}
    \label{tab:transformation_table}
    \vspace{-1.0em}
\end{table}

\section{Candidates Generation}
\label{sec:candidates_generation}
Next, in order to be able to verify whether an MT system correctly translated the property value in the source sentence, we automatically generate valid translation candidate sets for each property value. For some properties, such as number translation, we create exhaustive or near-exhaustive candidate sets. For other properties where the number of valid translations would be too big to do so, we instead create contrastive candidate pairs that demonstrate desired and undesired behavior. Note that candidate sets only need to be created once and can then be re-used for every tested system.

\subsection{Near-Exhaustive Candidate Sets}\label{sec:candidate-based-approach}
In this approach, we follow \citet{raunak-etal-2022-salted} in creating a set of all valid translations of each property value in the test (see example in \Cref{tab:transformation_table}). However, instead of manually designing candidate sets, we propose using the in-context learning \citep{gpt3} and multilingual capabilities of instruction-tuned LLMs \citep{wei2022finetuned} to accomplish the task. For each property value $\mathbf{x}_v$, we generate a set of translation candidates $\mathcal{C}^{\mathbf{x}_v}$ with ChatGPT (\texttt{gpt-3.5-turbo}) (see prompt\footnote{We use the same set of parameters as for the source sentence generation.} in \Cref{fig:prompt_candidates_no_format}). We tried to design demonstrations to encompass both correctness and completeness, including possible inflections. An example of demonstrations used for the currencies test can be seen in \Cref{apx:demonstrations_cand_sets}. Note that while we aim for completeness, i.e.\ all valid translations should be included in the candidate set, in practice we found that some rare translation choices may not be included in the automatically generated candidate sets. However, this will not impact pass-rates much because by nature rare translation choices appear in the MT system's output only in rare situations. In \Cref{sec:reliability} we perform a human assessment of the reliability of the generated candidate sets.

\subsection{Contrastive Candidate Pairs}
Some property values can span multiple words on the source side, potentially increasing the number of valid translations drastically. An example is idiomatic expressions, where there is an increased risk that the candidate set cannot exhaust all possibilities. 
To mitigate this issue, we propose using \emph{contrastive} candidate sets an alternative approach. 

Given a source property value, we generate a \textit{contrastive} candidate set $\mathcal{C}^{\mathbf{x}_v}_{\text{contra}}$ formed by a correct translation $c^{\mathbf{x}_v}_{\text{corr}}$, and a foil (incorrect) translation $c^{\mathbf{x}_v}_{\text{foil}}$. \Cref{apx:demonstrations_cand_sets_contrastive} shows an example prompt. Intuitively, an MT model should pass the test sentence if its translation is closer to $c^{\mathbf{x}_v}_{\text{corr}}$ than it is to $c^{\mathbf{x}_v}_{\text{foil}}$.

\begin{figure*}[!t]
\input{prompt_candidates_no_format.tex}
\caption{General template of the prompt used for generating near-exhaustive sets of candidate translations.}
\label{fig:prompt_candidates_no_format}
\end{figure*}

\begin{algorithm}[!t]
\DontPrintSemicolon
\KwIn{$\hat{\mathbf{y}}$: model translation; $c$: candidate translation; $e$: encoder}
\KwOut{$\text{max\_sim}({\hat{\mathbf{y}}, c})$}

$\text{max\_sim} \gets -\infty$\;
$n \gets |c|$\;
$\mathcal{G}_{\hat{\mathbf{y}}} \gets n\text{-gram}(\hat{\mathbf{y}},n)$\;
$\vc_{\text{emb}} \gets e(c)$\;
\For{$\mathbf{g} \in \mathcal{G}_{\hat{\mathbf{y}}}$}{
$\vg_{\text{emb}} \gets e(\mathbf{g})$\;
\If{$\text{sim}(\vg_{\text{emb}},\vc_{\text{emb}}) > \text{max\_sim}$}{
$\text{max\_sim} \gets \text{sim}(\vg_{\text{emb}},\vc_{\text{emb}})$\;
}
}

\Return $\text{max\_sim}$
\caption{Similarity score between translation and contrastive candidate.}
\label{alg:simlitter}
\end{algorithm}

\begin{figure*}[!t]
\begin{centering}
\includegraphics[width=0.55\textwidth]{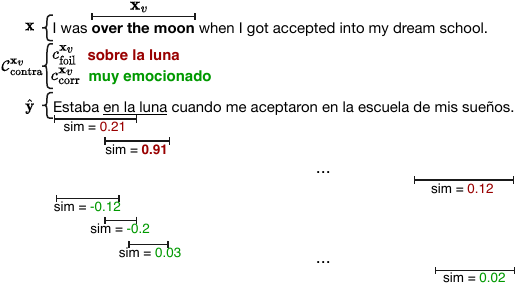}
	\caption{Example of the Contrastive Candidate Pairs approach, where \texttt{sim} indicates the semantic similarity between the correct candidate (`muy emocionado') and the $2$-grams (in green), and the foil candidate (`sobre la luna') and the $3$-grams of the MT translation (in red).}
	\label{fig:liter_idiom_approach}
	\end{centering}
\end{figure*}

\section{Pass-Fail Detector}
\label{sec:pass_fail_detector}

Equipped with these candidate sets, we now wish to mark every MT-translated sentence as either \textit{pass} or \textit{fail}. Depending on whether near-exhaustive or contrastive candidate pairs are used, we design pass-fail detectors based on string matching or semantic similarity, respectively.

As it is our goal to design tests that target specific language properties, our pass-fail detectors should only detect cases where the property value under the test is translated incorrectly. Unrelated translation errors should not cause a sentence to be marked as incorrect.\footnote{For our purposes, we do not consider whether the translated property is placed at the correct position in the target sentence, but only whether it is correct when considered in isolation. We argue that errors related to fluency and reordering are better evaluated through established accuracy-based metrics.}

\subsection{String Matching for Near-Exhaustive Candidate Sets}
For the near-exhaustive candidate sets, we define a pass-fail function $c(\hat{\mathbf{y}}, \mathcal{C}^{\mathbf{x}_v}) \in \{0,1\}$ that takes the model's translation $\hat{\mathbf{y}}$, and the candidates set $\mathcal{C}^{\mathbf{x}_v}$, and returns 1 (pass) if $\hat{\mathbf{y}}$ has a valid translation of the property value, i.e.\ if it has an element in $\mathcal{C}^{\mathbf{x}_v}$, and 0 (fail) otherwise:
\begin{equation}
c(\hat{\mathbf{y}}, \mathcal{C}^{\mathbf{x}_v}) = \begin{cases} 
      1 & \text{if}\;\; \hat{\mathbf{y}} \cap \mathcal{C}^{\mathbf{x}_v} \neq \emptyset\\
      0 & \text{otherwise.}
\end{cases}
\end{equation}

Specifically, we consider as pass an exact case-insensitive substring matching. Following Example \ref{ex:source_sentence}, where $\mathbf{x}_v = 4200.4$, if we are evaluating the En$\rightarrow$De decimal numbers translation capabilities of the model, we would consider the model passes the test if it outputs `4200,4', or `4.200,4'.\footnote{Note that this involves a design decision: The test case writers must make a decision whether or not the added decimal point is acceptable for their particular use cases.}

\subsection{Semantic Similarity for Contrastive Candidate Pairs}

For measuring the closeness of the property value translation to the contrastive candidates, we propose relying on the semantic similarity of word sequences representations extracted by a multilingual encoder \citep{reimers-gurevych-2019-sentence,reimers-gurevych-2020-making}.\footnote{Employing LLMs is also possible but not explored here because it needs to be applied for every evaluated MT system, incurring higher computational costs.} However, directly measuring the similarity between the translation of the property value and the candidate sets may be unreliable since they may differ in length and the location of the translation is unknown due to lack of word-level alignment. Instead, we propose that, for each candidate $c^{\mathbf{x}_v}_{\text{corr}}$ or $c^{\mathbf{x}_v}_{\text{foil}}$, we split the model's translation into $n$-grams, where $n$ is the number of words of the current candidate. Then, we measure the similarity between each of the $n$-grams and the candidates.

Given a translation and the \textit{contrastive} candidate set $\mathcal{C}^{\mathbf{x}_v}_{\text{contra}}$ formed by the correct and foil candidates, we define the pass-fail function as:
\begin{equation}
\resizebox{0.48\textwidth}{!}{$\displaystyle{
c(\hat{\mathbf{y}}, \mathcal{C}^{\mathbf{x}_v}_{\text{contra}}) = \begin{cases} 
      1 & \text{if}\;\;  \text{max\_sim}({\hat{\mathbf{y}}, c^{\mathbf{x}_v}_{\text{corr}}}) \geq \text{max\_sim}({\hat{\mathbf{y}}, c^{\mathbf{x}_v}_{\text{foil}}})\\
      0 & \text{otherwise.}
\end{cases}
}$}
\end{equation}

\Cref{alg:simlitter} formalizes the computation of $\text{max\_sim}$ function, \Cref{fig:liter_idiom_approach} shows an example.

\begin{table*}[!t]
\resizebox{\linewidth}{!}{
\begin{tabular}{cccccccccc}
\toprule
\multicolumn{1}{c}{\multirow{2}{*}{\textbf{Model}}} & \multicolumn{3}{c}{\textbf{En$\rightarrow$De}} &                        \multicolumn{3}{c}{\textbf{En$\rightarrow$Es}}                                                    & \multicolumn{3}{c}{\textbf{En$\rightarrow$Ja}}                              \\ \cmidrule{2-10} 
\multicolumn{1}{c}{}                       & spBLEU & ChrF  & COMET-22 & spBLEU                     & ChrF                     & COMET-22                     & spBLEU                & ChrF                  & COMET-22              \\ \cmidrule{1-10} 
M2M 418M                                   & 31.08  & 57.22 & 79.49    & 25.33                      & 51.26                    & 80.63                        & 23.57                 & 32.22                 & 84.84                 \\
M2M 1.2B                                   & 39.37  & 62.51 & 85.35    & 29.06                      & 53.85                    & 84.22                        & 27.46                 & 35.25                 & 87.63                 \\
NLLB 600M                                  & 38.88  & 61.85 & 85.89    & 30.65                      & 54.76                    & 85.34                        & 18.75                 & 29.62                 & 86.72                 \\
NLLB 3.3B                                  & 44.41  & 65.26 & 87.98    & 32.69                      & 56.09                    & 86.39                        & 20.76                 & 32.5                  & 88.12                 \\
OPUS MT (Bil)                              & 40.96  & 63.49 & 84.61    & 30.57                      & 54.97                    & 84.9                         &  -                     &  -                     &   -                    \\
WMT21 (En-X)                               & \textbf{49.38}  & \textbf{68.94} & 88.76    & -                          & -                        & -                            & 39.89                 & 44.95                 & 91.95                     \\
Commercial system     & 49.34  & 68.84 & \textbf{89.34}    & \textbf{34.43}                      & \textbf{57.58}                    & \textbf{86.92}                        & \textbf{41.05}                 &\textbf{47.06}                 & \textbf{92.19}\\ \bottomrule
\end{tabular}
}
\caption{Translation scores of the different models used in FLORES-200 devtest set.}
\label{tab:flores_results}
\end{table*}

\section{Evaluation Metrics}

Having established pass-fail detection for individual sentences, the final step is to compute aggregated \textit{pass rates} across test sets. Appealingly, pass rates are naturally expressed as percentages, making them intuitive to interpret.

\subsection{Macro Pass Rate}
Let us assume that we have computed pass-fail results across a behavioral test set consisting of $N$ test cases (sentences). From a statistical viewpoint, we have access to a sample ${\mathcal{X}=\{c(\hat{\mathbf{y}}^n, \mathcal{C}^{\mathbf{x}^n_v})\}_{n=1}^{N}}$, drawn from some unknown distribution over test cases, $F$.
The expectation of the true pass rate can be computed as follows:
\begin{equation}
\text{PR}^{(\mathcal{X})} = \frac{1}{N}\sum_{n}^{N} c(\hat{\mathbf{y}}^n, \mathcal{C}^{\mathbf{x}^n_v})
\end{equation}

One issue that arises in practice is that property values themselves follow a long tail pattern: Certain values appear relatively frequently, while many other values appear only once across the generated test set. This can make pass rates overly sensitive to whether models happen to perform well for these particular values. To mitigate this issue, we assume a generative story in which property values are drawn from a uniform distribution, and consequently compute the expected pass rate as the macro average across property values:
\begin{equation}
\text{MPR}^{(\mathcal{X})} = \frac{1}{|\mathcal{V}|} \sum_{v \in \mathcal{V}} \frac{1}{N_{v}}\sum_{i}^{N_{v}} c(\hat{\mathbf{y}}^i, \mathcal{C}^{\mathbf{x}^i_v})
\end{equation}
where $\mathcal{V}$ refers to the set of distinct property values, and $N_{v}$ to the number of examples associated with each specific property value.
\subsection{Confidence Intervals}
Although previous work performing behavioral testing for MT shows point estimate scores, confidence intervals provide a more reliable approach to statistical analysis, as they quantify the uncertainty associated with that estimate, and ensure the sample size is large enough.
To compute confidence intervals for our estimator $\text{MPR}$ we use the Bootstrap method \citep{bootstrap}, which performs sampling with replacement from $\mathcal{X}$, generating $K$ resamples $\{\mathcal{Y}^1, \cdots, \mathcal{Y}^K\}$, from which we compute their corresponding macro pass rates $\{\text{MPR}^{(\mathcal{Y}^{1})}, \cdots, \text{MPR}^{(\mathcal{Y}^{K})}\}$ to construct the bootstrap distribution  $\text{MPR}_{\text{boot}}$. Assuming the distribution of $\mathcal{X}$ is a reasonable approximation of the population distribution $F$, confidence intervals can be derived from $\text{MPR}_{\text{boot}}$. For that purpose, we compute the percentile bootstrap interval for $\alpha=0.05$ provided by \textsc{SciPy} library \citep{2020SciPy-NMeth}.
\subsection{Paired Bootstrap}\label{sec:paired_bootstrap}
The paired bootstrap is a statistical resampling technique used to assess the uncertainty and make inferences about the difference between two samples. Paired bootstrap allows us to compare the property's sample of passes/fails for two different models \citep{koehn-2004-statistical}. By following the resampling process outlined in the previous section, if a model consistently outperforms the other in 95\% of the iterations, we can assert with 95\% statistical significance that it is superior.

\section{Properties to Test}
We design a number of tests and use our proposed framework to evaluate MT models in multiple properties. The chosen properties, also studied in the literature~\citep{wang-etal-2021-easy,raunak-etal-2022-salted}, have two important qualities that make them useful for evaluating translation systems: vital for producing high-quality translations, yet posing a challenge when assessing through conventional evaluation metrics.

\paragraph{Numbers.} We conduct independent assessments for integers (e.g. 1887), decimals (e.g. 154.32), and large numbers (e.g. 200 billion). Large numbers have the format ``integer/decimal million/billion/trillion''. We create near-exhaustive candidate sets of valid number translations and check if the translation matches any candidate.
\paragraph{Physical Units.} We build near-exhaustive candidate sets for evaluating the translations of diverse units including those related to weight, length, time, or temperature \textit{inter alia} (e.g. inches). Translations are evaluated by string matching.
\paragraph{Emojis, Names, and Web Terms.} Via string matching we check whether the translated text retains the same property instantiation found in the source text. Candidate sets for these tests are thus considered to be exhaustive.
\paragraph{Currencies.} We consider currencies appearing in the ISO code format (e.g. EUR). Near exhaustive candidate sets are built allowing translations into the same ISO code, variations of the currency name or its symbol (e.g. for En$\rightarrow$Es: EUR/euro/euros/€), then a string matching pass-fail detection is employed.
\paragraph{Idioms.} Idiomatic expressions pose significant challenges for MT systems due to their non-literal nature and potential large sequence length. We use idioms as a test bench for the use of contrastive candidate pairs (incorrect literal translation candidate vs. correct meaning translation) and semantic similarity detection procedure.

\begin{figure*}[!t]\begin{centering}\includegraphics[width=0.87\textwidth]{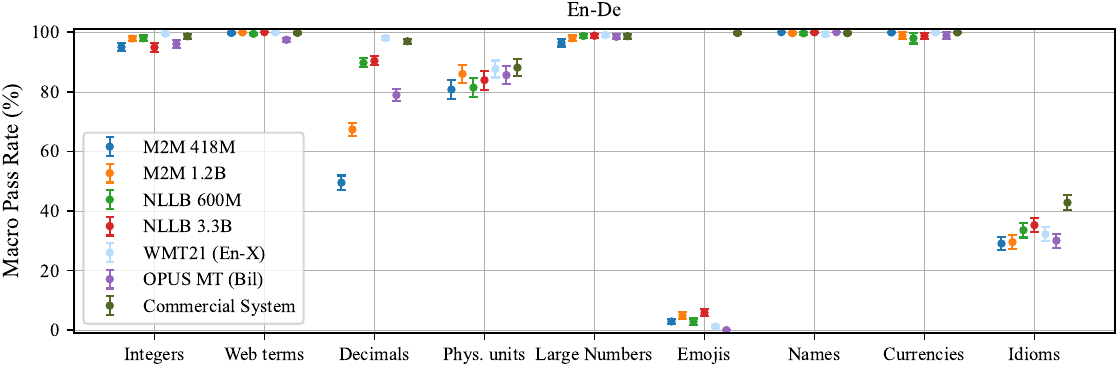}
	\caption{En$\rightarrow$De macro pass rates and confidence intervals across tested systems.}
	\label{fig:en_de_ci_1}
	\end{centering}
\end{figure*}

\section{Models Comparison}
In this section, we introduce the tested models and present results obtained via standard metrics as well as our proposed framework.
\subsection{Experimental Setup}
We test widely-used open-source MT models, as well as a commercial system. We aim to select models that perform very strongly, while also differing in some important aspects (e.g.\ bilingual vs.\ multilingual).

In the multilingual domain, we experiment with the 600M and the 3.3B parameters models of No Language Left Behind project (NLLB)~\citep{nllb2022}, and the Many-to-Many (MLM-100) family of Multilingual models~\citep{m2m100} (418M and 1.2B parameters models). Additionally, we evaluate the WMT21: multilingual (7 En$\rightarrow$X directions) 4.7B dense model~\citep{tran-etal-2021-facebook}, part of Meta's WMT-21 News Translation task participation~\citep{wmt-2021-machine}. We also assess OPUS-MT~\citep{tiedemann-thottingal-2020-opus} En$\rightarrow$Es and En$\rightarrow$De bilingual models trained on OPUS dataset~\citep{tiedemann-2012-parallel}. Lastly, we included results from an anonymous commercial system.

Besides our proposed metrics, we also evaluate the models on FLORES-200~\citep{nllb2022} in En$\rightarrow$De, En$\rightarrow$Es, and En$\rightarrow$Ja via string-based metrics spBLEU\footnote{\texttt{SACREBLEU} signature: \texttt{nrefs:1|case:mixed|eff:no|\\tok:flores101|smooth:exp|version:2.3.1}}~\citep{papineni-etal-2002-bleu} and ChrF\footnote{\texttt{SACREBLEU} signature: \texttt{nrefs:1|case:mixed|eff:yes|\\nc:6|nw:0|space:no|version:2.3.1}}~\citep{popovic-2015-chrf} as implemented in \textsc{Sacrebleu}~\citep{post-2018-call}, as well as the neural-based metric COMET-22\footnote{\texttt{Unbabel/wmt22-comet-da}} \citep{rei-etal-2020-comet}.

\begin{table}[t!]
    \centering
    \footnotesize
    \begin{tabular}{@{}p{1.2cm}p{2.6cm}p{2.6cm}}
      \toprule
      \multicolumn{1}{c}{\textbf{Model}} &\multicolumn{1}{c}{\textbf{Source Sentence}} & \multicolumn{1}{c}{\textbf{Translation}} \\
      \midrule
      OPUS MT (Bil) & The article I read on \nolinkurl{www.scientificjournal.org} was very informative. & El artículo que leí en \textcolor{red}{\nolinkurl{www.cientificojournal.org}} fue muy informativo.\\\midrule
       Commercial system & ... our town's population was counted as 12,577. & ... población de nuestra ciudad se contabilizó en 12\textcolor{red}{,}577.\\
      \bottomrule
    \end{tabular}
    \caption{Examples flagged as failed translations.}
    \label{tab:examples_fails}
\end{table}

\subsection{General Translation Accuracy}

We first measure general translation performance across language pairs for standard reference-based metrics (\Cref{tab:flores_results}). The commercial system performs best across the board, followed by the WMT21 model. In the following sections, we dive deeper into the different capabilities. 
\subsection{Behavioral Tests Results}\label{sec:beh_test_res}
As an illustrative example, macro pass rate confidence intervals across property types and models for the En$\rightarrow$De direction are presented in \Cref{fig:en_de_ci_1}. The complete results can be found in \Cref{sec:pr_conf_inte}.

\paragraph{Commercial system is most consistent across properties.} 
This is especially true for emoji translations, where open-source models lack most emojis in their vocabulary. However, it is noteworthy that its performance is subpar in the context of En$\rightarrow$Es integers and En$\rightarrow$Ja large numbers. After manual inspection (see examples in \Cref{tab:examples_fails}), we attribute the lower integers translation performance to the fact that it uses the comma as the thousands separator. Note that this behavior can be acceptable depending on the country; behavioral tests must be designed to reflect the intended behavior.

\begin{figure*}[t!]
\begin{center}
\includegraphics[width=0.87\textwidth]{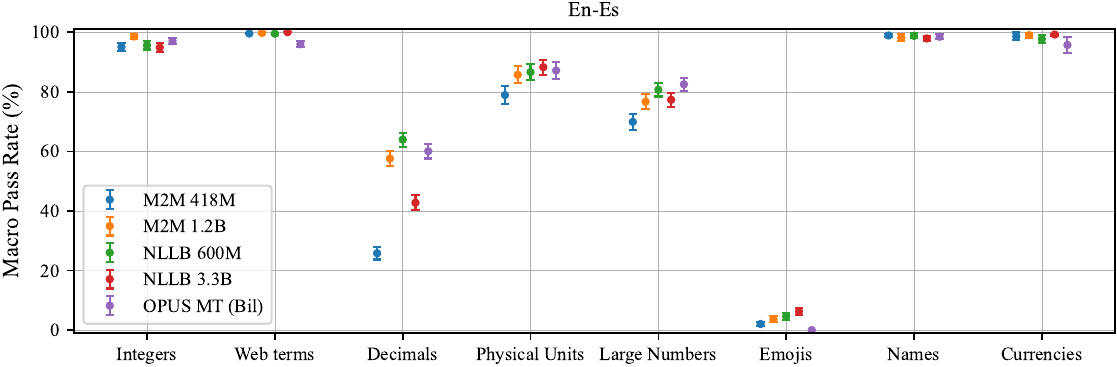}
\includegraphics[width=0.87\textwidth]{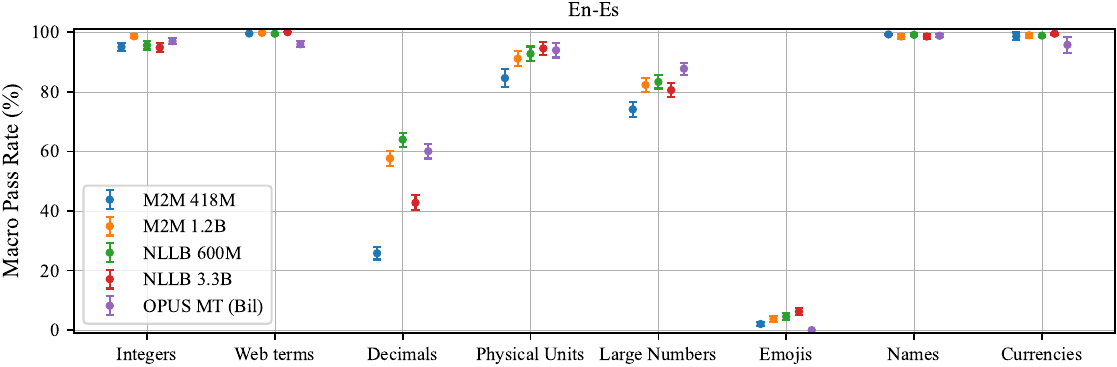}
\includegraphics[width=0.87\textwidth]{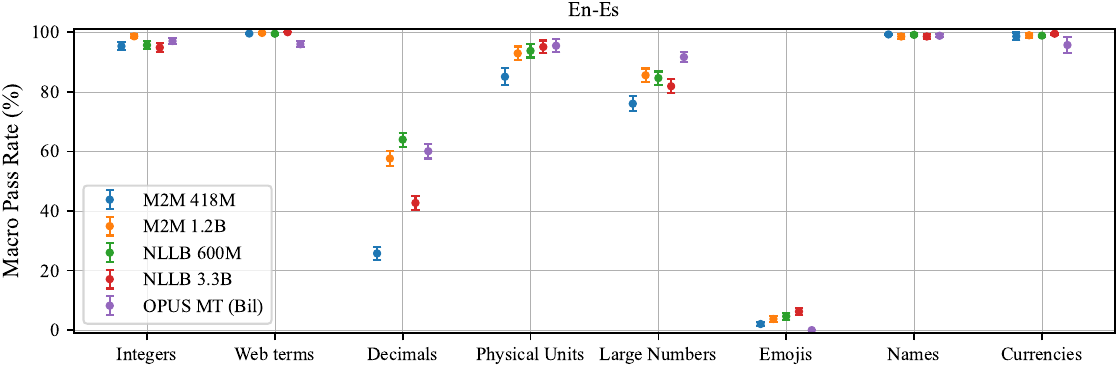}
\caption{From top to bottom, En$\rightarrow$Es Confidence Intervals after each annotation iteration (see \Cref{sec:reliability}).}
\label{fig:ci_annotations}
\end{center}
\end{figure*}

\paragraph{Bilingual models struggle with web terms.}
Although the multilingual models mostly manage to preserve web terms without alteration, both tested bilingual models (for En$\rightarrow$Es and En$\rightarrow$De) underperform in that property (see \Cref{fig:en_de_ci_1} and \Cref{fig:ci_annotations} top). Most fail cases contain Spanish words inside the translated web terms (\Cref{tab:examples_fails}). We hypothesize that this occurs because they are trained to exclusively translate into Spanish, which consequently hinders their ability to generate content in other languages, and is therefore an intrinsic limitation of bilingual models.

\paragraph{Scaling models help increase capabilities.} In most of the settings, scaling the model of the same family shows increased performance, for instance, physical units and idioms in \Cref{fig:en_de_ci_1}. However, there are some counter-examples, like in the case of En$\rightarrow$Ja decimals and integers tests.

\paragraph{WMT21 is the strongest open-source model.} The WMT21 model consistently exhibits superior performance compared to other open-source models in both En$\rightarrow$De and En$\rightarrow$Ja tests. In \Cref{tab:en_de_integers_paired} we show how paired bootstrap enables model comparison, revealing that WMT21 outperforms other models in the integers En$\rightarrow$De test.

\paragraph{Idioms.} Results for the Idioms property test are presented in \Cref{apx:idioms_ci}. The ability to translate idioms is generally low (i.e.\ overly literal), in accordance with recent findings \citep{dankers-etal-2022-transformer}. It is worth noting that results are similar in the three language directions, with the commercial system and NLLB 3.3B showing comparable performance.

\begin{table}[t]
\resizebox{0.49\textwidth}{!}{
\begin{tabular}{cccc}
\toprule
\textbf{Model A} & \textbf{Model B} & \textbf{Winner} & \textbf{p-value} \\ \midrule
M2M 418M      & M2M 1.2B      & M2M 1.2B      & 0.0   \\
M2M 418M      & NLLB 600M     & NLLB 600M     & 0.0   \\
M2M 418M      & NLLB 3.3B     & M2M 418M      & 0.476 \\
M2M 418M      & WMT21 (En-X)  & \underline{WMT21 (En-X)}  & 0.0   \\
M2M 418M      & OPUS MT (Bil) & OPUS MT (Bil) & 0.106 \\
M2M 418M      & Commercial system        & Commercial system        & 0.0   \\
M2M 1.2B      & NLLB 600M     & NLLB 600M     & 0.461 \\
M2M 1.2B      & NLLB 3.3B     & M2M 1.2B      & 0.0   \\
M2M 1.2B      & WMT21 (En-X)  & \underline{WMT21 (En-X)}  & 0.001 \\
M2M 1.2B      & OPUS MT (Bil) & M2M 1.2B      & 0.004 \\
M2M 1.2B      & Commercial system        & Commercial system        & 0.102 \\
NLLB 600M     & NLLB 3.3B     & NLLB 600M     & 0.0   \\
NLLB 600M     & WMT21 (En-X)  & \underline{WMT21 (En-X)}  & 0.003 \\
NLLB 600M     & OPUS MT (Bil) & NLLB 600M     & 0.005 \\
NLLB 600M     & Commercial system        & Commercial system        & 0.142 \\
NLLB 3.3B     & WMT21 (En-X)  & \underline{WMT21 (En-X)}  & 0.0   \\
NLLB 3.3B     & OPUS MT (Bil) & OPUS MT (Bil) & 0.118 \\
NLLB 3.3B     & Commercial system        & Commercial system        & 0.0   \\
WMT21 (En-X)  & OPUS MT (Bil) & \underline{WMT21 (En-X)}  & 0.0   \\
WMT21 (En-X)  & Commercial system        & \underline{WMT21 (En-X)}  & 0.024 \\
OPUS MT (Bil) & Commercial system        & Commercial system        & 0.0 \\ \bottomrule
\end{tabular}
}
\caption{Paired Bootstrap En$\rightarrow$De Integers test results. We make a 95\% statistically significant conclusion that the WMT21 system is better than the rest of the models.}
\label{tab:en_de_integers_paired}
\end{table}

\section{Reliability of the Proposed Approach}
\label{sec:reliability}
To assess the reliability of the proposed approach, in this section we analyze the robustness of source sentence generation and pass-fail detection.

\subsection{Analysis of Source Sentence Generation}\label{sec:analysis_source}
One potential concern with the proposed method is whether the generated source sentences are diverse enough and do not become repetitive after a few rounds of generation.\footnote{The \textit{naturalness} of outputs, another potential concern, has been extensively dealt with elsewhere \citep{ouyang2022training}.}
 A standard method for quantifying the diversity in a corpus is \textit{distinct $n$-grams} \citep{li-etal-2016-diversity}, which computes the ratio of unique $n$-grams to the total number of $n$-grams present. In our case, we are interested in assessing the diversity of each generated source sentence compared to the previous generations. To that end, we propose a metric to measure this aspect. Given the set of unique $n$-grams generated up to sentence $\mathbf{x}_t$ ($\mathcal{G}^{n}_{\mathbf{x}_{<t}}$), we measure the proportion of unique $n$-grams in each newly generated sentence ($\mathcal{G}^{n}_{\mathbf{x}_t}$) that are not present in $\mathcal{G}^{n}_{\mathbf{x}_{<t}}$:
\begin{equation}
\text{div}_{n} (\mathbf{x}_t) = \frac{\mathcal{G}^{n}_{\mathbf{x}_t} \setminus \mathcal{G}^{n}_{\mathbf{x}_{<t}}}{\mathcal{G}^{n}_{\mathbf{x}_t}}
\end{equation}

\begin{figure}[!t]
\begin{centering}
\includegraphics[width=0.49\textwidth]{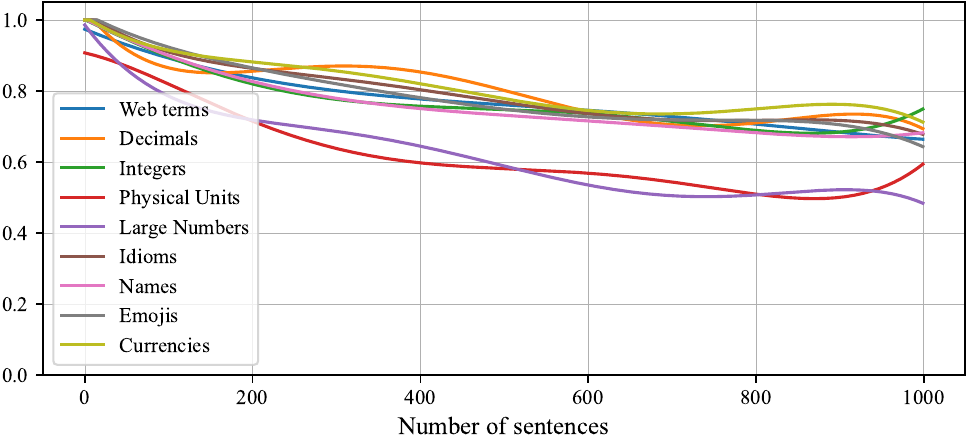}
	\end{centering}
    \caption{3-gram diversity scores ($\text{div}_{3} (\mathbf{x}_t)$) along generation steps across different properties.}
    \label{fig:src_generation_diversity}
\end{figure}

\Cref{fig:src_generation_diversity} shows 3-gram diversity along 1000 generated sentences after fitting a polynomial regression.  We observe that the diversity drop is mild even after 500 sentences, where for most of the tests, 60\% of newly generated 3-grams are novel.

Furthermore, we observe that the sentence generator produces sentences that comply with instructions, indicated by the high proportion of the original sentences that pass filtering. In the majority of cases, over 70\% of the LLM-generated sentences successfully pass the filtering steps outlined in \Cref{sec:source_sent_generation}, as seen in \Cref{tab:src_generation_efficiency} (middle column). The right column shows the percentage of unique values, which naturally vary strongly depending on the property.

\begin{table}[!t]
\begin{center}
\resizebox{0.9\columnwidth}{!}{%
\begin{tabular}{lcc}
\toprule
\textbf{Property}        & \textbf{Sentences kept}  & \textbf{Unique values}  \\ \midrule
Web Terms         & 79.3{\%} & 92.5{\%} \\
Decimals           & 74.1{\%} & 76.9{\%} \\
Integers           & 62.1{\%} & 39.3{\%} \\
Physical Units    & 83.3{\%} & 15.9{\%} \\
Large Numbers      & 66.1{\%} & 37.1{\%} \\
Idioms            & 83.8{\%} & 69.0{\%} \\
Names          & 86.1{\%} & 17.9{\%} \\
Emojis             & 88.5{\%} & 29.7{\%} \\
Currencies          & 66.8{\%} & 5.2{\%} \\
\bottomrule
\end{tabular}
}
\end{center}
    \caption{Percentage of source sentences that pass filtering, and percentage of filtered sentences that introduce a new property value.}
    \label{tab:src_generation_efficiency}
    \vspace{-1.0em}
\end{table}

\begin{figure}[!t]\begin{centering}\includegraphics[width=0.49\textwidth]{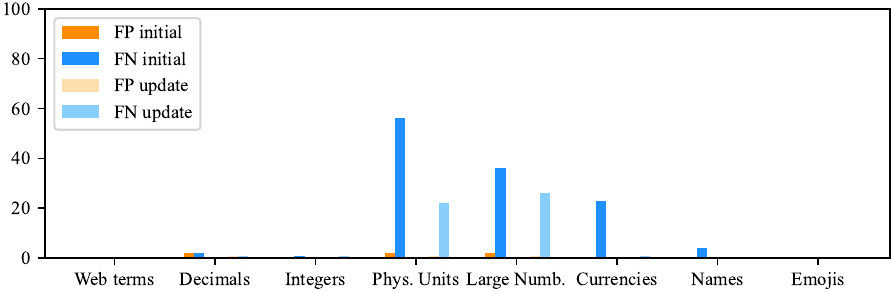}
	\caption{Error rates detected in two rounds of annotations on En$\rightarrow$Es.}
	\label{fig:error_rates_updates}
	\end{centering}
\end{figure}

\subsection{Analysis of Pass-Fail Detection}
The reliability of the proposed pass-fail detection depends mainly on whether candidate sets are (1) complete and (2) do not contain wrong candidates.

We analyze this by sampling 100 random test cases that were marked as \textit{pass} (positives), and another 100 examples marked as \textit{fail} (negatives). We manually annotate whether test results were correct or incorrect. \Cref{fig:error_rates_updates} shows false positives and false negatives (FP initial and FN initial). We observe that while for most properties these were low, for some test cases (namely physical units, large numbers, currencies) there were a significant number of FNs, which would lead to underestimated pass rates. We argue that erring on the side of FNs is generally preferable, because it prevents us from overestimating the strength of models, and because it would trigger a debugging effort which would quickly surface issues stemming from FNs.

To obtain more accurate pass-rates for all properties, we can manually remove candidates causing a FP and add missing candidates producing a FN. We do this for the test cases analyzed above, and then draw another random sample from both \textit{pass} and \textit{fail} categories. \Cref{fig:error_rates_updates} shows that the updated FPs and FNs are now negligible.

While in our experience, human intervention as outlined above is only a minor effort, the issue remains as to whether systems can be compared to one another without the need for human intervention, even in the presence of existing FNs. To understand this better, we plot macro pass rates with confidence intervals across annotation iterations in \Cref{fig:ci_annotations}. As expected, for physical units, large numbers, and currencies, pass rates move upwards. However, the effect is general across models, suggesting that relative ordering between models can be reasonably approximated in the initial attempt, i.e.\ without human intervention.

In addition, we assess the pass-fail detection of idioms. Given that the decision is made via semantic similarity for contrasting pairs, addressing issues in the candidate sets is more challenging. Consequently, we conducted a single evaluation iteration with 100 \textit{pass}/\textit{fail} examples, respectively, on two language pairs. 
For En$\rightarrow$De, we observed 59 FPs / 16 FNs; En$\rightarrow$Es had 50 FPs / 11 FNs. We hypothesize high FPs are caused by idiom and its figurative meaning being present within the source sentence, interfering with the $n$-grams comparison. We leave further investigation for future research.

\section{Related Work}
Recent works have applied behavioral testing for evaluating machine translation systems. \citet{wang-etal-2021-easy} designed tests for numerical translation capabilities by relying on fixed templates for source sentence generation. \citet{raunak-etal-2022-salted} proposed SALTED, a set of manually designed error detectors that are applied to millions of sentences from standard datasets. Beyond behavioral testing, a large number of \textit{challenge sets} have been developed for machine translation \cite{popovic-castilho-2019-challenge}. Although useful, most of these evaluation tools require major human efforts for creation, evaluation, or expanding to other languages. Although there have been attempts to automatize the creation of behavioral tests \citep{yang-etal-2022-testaug}, this has been limited to simple NLP tasks.

Our work also relates to the use of LLMs as evaluators for Machine Translation systems \citep{kocmi2023large}, as well as for text generation in a broader sense \citep{liu2023geval,xu2023instructscore}, which extend the growing body of research on multi-dimensional text generation evaluation \citep{zhong-etal-2022-towards, bartscore}.

Behavioral testing aims to evaluate the behavior of systems under realistic conditions, contrasting it from the literature on adversarial data generation \citep{belinkov2018synthetic,zhang-etal-2021-crafting}.

\section{Conclusions}
In this work, we have presented a method that automates the creation of behavioral tests to perform fine-grained evaluation of MT systems capabilities. We use Large Language Models to generate source sentences composed of fragments of specific language properties (integers, web terms, etc.), as well as translations of these properties. For property types formed by multiple words, we further extend the proposed method into a contrastive setting and show its usefulness in evaluating idiomatic expressions. To the best of our knowledge, our research represents the first attempt to develop MT behavioral tests by leveraging LLMs. Finally, we apply the proposed framework to evaluate open-source models on three language pairs.


\bibliography{anthology,custom}
\bibliographystyle{acl_natbib}

\newpage
\appendix
\onecolumn
\clearpage
\section{Limitations}
While the proposed evaluation framework seeks to address a broad spectrum of languages, the experiments conducted in this study are limited to three language pairs. Due to its reliance on the capacity of LLMs to produce high-quality candidate translations, we cannot guarantee accurate results when applied to language pairs involving a low-resource language using current LLMs. Moreover, the method is designed to work only on properties that appear as a continuous chunk of text in both source and target languages and are not scattered across a sentence.
\clearpage
\section{Example of Demonstrations for Exhaustive Candidate Set Generation}\label{apx:demonstrations_cand_sets}
\noindent
\begin{minipage}{\textwidth}
        \centering
        \vspace{3ex}
\input{prompt_candidates_no_format_currency}
\captionof{figure}{General template of the prompt used for generating a set of candidate translations.}
\label{fig:prompt_candidates_no_format_currency}
\end{minipage}

\section{Example of Demonstrations for Contrastive Candidate Pairs Generation}\label{apx:demonstrations_cand_sets_contrastive}

\noindent\begin{minipage}{\textwidth}
        \centering
        \vspace{3ex}
\input{prompt_candidates_contrastive_no_format.tex}
\label{fig:prompt_candidates_contrastive_idioms}
\captionof{figure}{Prompt used for generating contrastive candidate pairs for the case of idioms. For the literal translation (foil) we prompt ChatGPT with the idiom in isolation. Conversely, in order to facilitate the `understanding' of the idiom's figurative connotation, for generating correct candidates we present it within the full sentence.}
\end{minipage}

\begin{figure*}[!t]

\end{figure*}

\clearpage
\section{Idioms Test Results}\label{apx:idioms_ci}
\noindent\begin{minipage}{\textwidth}
        \centering
        \vspace{3ex}
\includegraphics[width=0.7\textwidth]{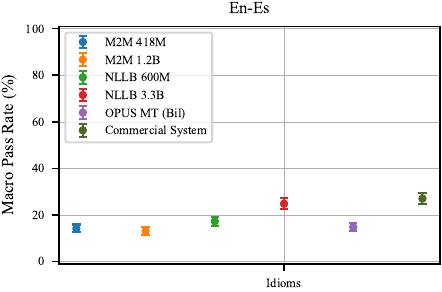}
\includegraphics[width=0.7\textwidth]{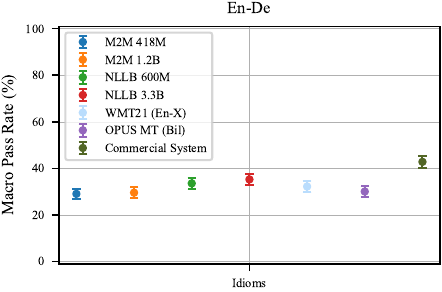}
\includegraphics[width=0.7\textwidth]{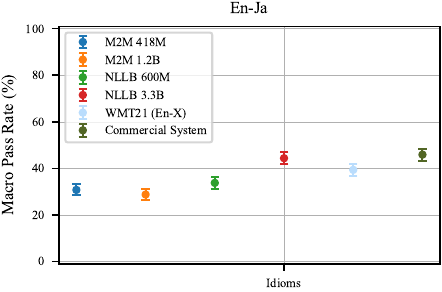}

 \captionof{figure}{Idioms Test Confidence Intervals across language pairs.}
	\label{fig:pass_rate_idioms_chatgpt}
\end{minipage}

\clearpage
\section{Pass Rate Confidence Intervals}\label{sec:pr_conf_inte}
\vspace{3ex}

\noindent\begin{minipage}{\textwidth}
        \centering
        \vspace{3ex}
\includegraphics[width=0.99\textwidth]{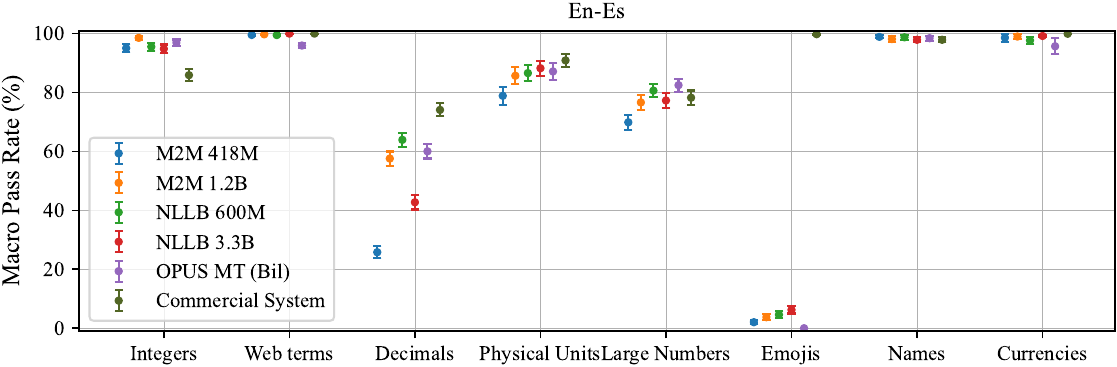}
 \captionof{figure}{Macro pass rate confidence intervals for En$\rightarrow$Es tests.}
\label{fig:OS_en_es_MPR_rdmv2_no_idioms.png}
\end{minipage}

\vspace{5ex}
\noindent\begin{minipage}{\textwidth}
\centering
\renewcommand{\arraystretch}{1.5}
\resizebox{\linewidth}{!}{
\begin{tabular}{ccccccc}
\toprule
\multirow{2}{*}{\textbf{Property}}                      & \multicolumn{5}{c}{\textbf{Macro Pass Rate} (\%)}         \\ \cmidrule{2-7} 
               & \textbf{M2M 418M}  & \textbf{M2M 1.2B}  & \textbf{NLLB 600M} & \textbf{NLLB 3.3B} & \textbf{OPUS MT (Bil)} & \textbf{Commercial system}            \\ \cmidrule{1-7} 
Web terms      & {[}0.993, 0.998{]} & {[}0.995, 1.0{]}   & {[}0.992, 0.998{]} & {[}1.0, 1.0{]}     & {[}0.95, 0.97{]}   & {[}1.0, 1.0{]}     \\
Decimals       & {[}0.236, 0.278{]} & {[}0.551, 0.602{]} & {[}0.615, 0.665{]} & {[}0.401, 0.451{]} & {[}0.576, 0.625{]} & {[}0.719, 0.764{]} \\
Integers       & {[}0.939, 0.966{]} & {[}0.979, 0.993{]} & {[}0.943, 0.969{]} & {[}0.935, 0.963{]} & {[}0.959, 0.98{]}  & {[}0.838, 0.88{]}  \\
Physical Units & {[}0.824, 0.879{]} & {[}0.905, 0.952{]} & {[}0.915, 0.96{]}  & {[}0.93, 0.972{]}  & {[}0.933, 0.975{]} & {[}0.952, 0.987{]} \\
Large Numbers  & {[}0.736, 0.787{]} & {[}0.834, 0.878{]} & {[}0.824, 0.868{]} & {[}0.795, 0.842{]} & {[}0.898, 0.935{]} & {[}0.863, 0.907{]} \\
Emojis         & {[}0.014, 0.027{]} & {[}0.027, 0.048{]} & {[}0.035, 0.058{]} & {[}0.05, 0.075{]}  & {[}0.0, 0.0{]}     & {[}0.996, 1.0{]}   \\
Names          & {[}0.991, 0.994{]} & {[}0.977, 0.994{]} & {[}0.986, 0.996{]} & {[}0.978, 0.993{]} & {[}0.983, 0.993{]} & {[}0.978, 0.993{]} \\
Currencies     & {[}0.973, 0.999{]} & {[}0.982, 0.997{]} & {[}0.985, 0.992{]} & {[}0.992, 0.998{]} & {[}0.93, 0.985{]}  & {[}0.999, 1.0{]}   \\
Idioms         & {[}0.125, 0.159{]} & {[}0.114, 0.148{]} & {[}0.153, 0.192{]} & {[}0.225, 0.271{]} & {[}0.129, 0.166{]} & {[}0.247, 0.294{]}  \\  \bottomrule
\end{tabular}
}
\captionof{figure}{Macro pass rate confidence intervals for En$\rightarrow$Es tests.}
\label{tab:conf_intervals_en_es}
\end{minipage}

\noindent\begin{minipage}{\textwidth}
        \centering
        \vspace{3ex}
\includegraphics[width=0.99\textwidth]{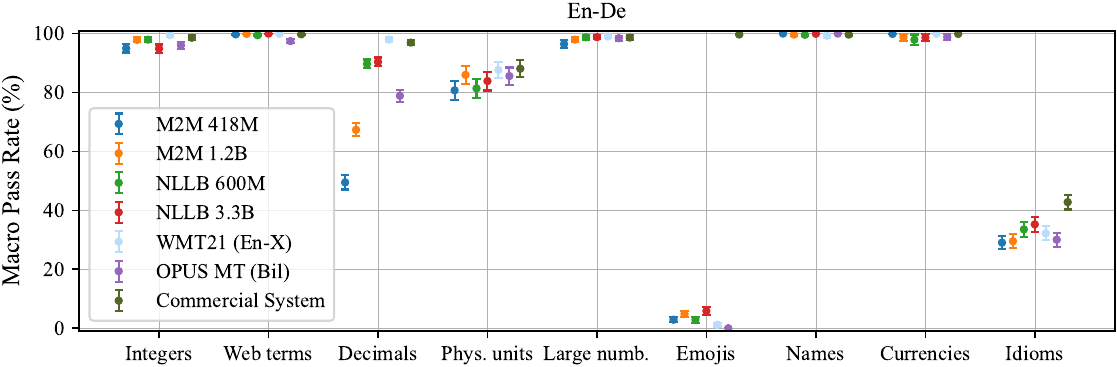}
 \captionof{figure}{Macro pass rate confidence intervals for En$\rightarrow$De tests.}
\label{fig:en_de_MPR.png}
\end{minipage}

\clearpage

\vspace{5ex}
\noindent\begin{minipage}{\textwidth}
\centering
\renewcommand{\arraystretch}{1.5}
\resizebox{\linewidth}{!}{
\begin{tabular}{cccccccc}
\toprule
\multirow{2}{*}{\textbf{Property}}                      & \multicolumn{5}{c}{\textbf{Macro Pass Rate} (\%)}         \\ \cmidrule{2-8} 
               & \textbf{M2M 418M}  & \textbf{M2M 1.2B}  & \textbf{NLLB 600M} & \textbf{NLLB 3.3B} & \textbf{WMT21 (En-X)} & \textbf{OPUS MT (Bil)} & \textbf{Commercial system}            \\ \cmidrule{1-8} 
Web terms      & {[}0.995, 1.0{]}   & {[}0.998, 1.0{]}   & {[}0.991, 0.998{]} & {[}1.0, 1.0{]}     & {[}0.998, 1.0{]}   & {[}0.967, 0.982{]} & {[}0.995, 1.0{]}   \\
Decimals       & {[}0.471, 0.519{]} & {[}0.651, 0.696{]} & {[}0.882, 0.912{]} & {[}0.889, 0.919{]} & {[}0.973, 0.987{]} & {[}0.769, 0.809{]} & {[}0.961, 0.978{]} \\
Integers       & {[}0.936, 0.964{]} & {[}0.97, 0.987{]}  & {[}0.97, 0.989{]}  & {[}0.935, 0.963{]} & {[}0.99, 1.0{]}    & {[}0.948, 0.973{]} & {[}0.978, 0.994{]} \\
Physical Units & {[}0.775, 0.84{]}  & {[}0.83, 0.89{]}   & {[}0.781, 0.847{]} & {[}0.807, 0.871{]} & {[}0.848, 0.905{]} & {[}0.827, 0.885{]} & {[}0.853, 0.909{]} \\
Large Numbers  & {[}0.952, 0.977{]} & {[}0.97, 0.989{]}  & {[}0.98, 0.995{]}  & {[}0.981, 0.995{]} & {[}0.984, 0.997{]} & {[}0.976, 0.993{]} & {[}0.978, 0.995{]} \\
Emojis         & {[}0.02, 0.038{]}  & {[}0.037, 0.06{]}  & {[}0.018, 0.039{]} & {[}0.046, 0.071{]} & {[}0.006, 0.018{]} & {[}0.0, 0.0{]}     & {[}0.994, 1.0{]}   \\
Names          & {[}1.0, 1.0{]}     & {[}0.993, 1.0{]}   & {[}0.992, 1.0{]}   & {[}0.999, 1.0{]}   & {[}0.986, 1.0{]}   & {[}1.0, 1.0{]}     & {[}0.993, 1.0{]}   \\
Currencies     & {[}0.998, 1.0{]}   & {[}0.976, 1.0{]}   & {[}0.962, 0.997{]} & {[}0.976, 0.998{]} & {[}0.999, 1.0{]}   & {[}0.977, 1.0{]}   & {[}0.998, 1.0{]}   \\
Idioms         & {[}0.268, 0.313{]} & {[}0.272, 0.319{]} & {[}0.311, 0.36{]}  & {[}0.328, 0.377{]} & {[}0.298, 0.346{]} & {[}0.277, 0.324{]} & {[}0.403, 0.453{]}  \\  \bottomrule
\end{tabular}
}
\captionof{figure}{Macro pass rate confidence intervals for En$\rightarrow$De tests.}
\label{tab:conf_intervals_en_de}
\end{minipage}

\noindent\begin{minipage}{\textwidth}
        \centering
        \vspace{3ex}
\includegraphics[width=0.99\textwidth]{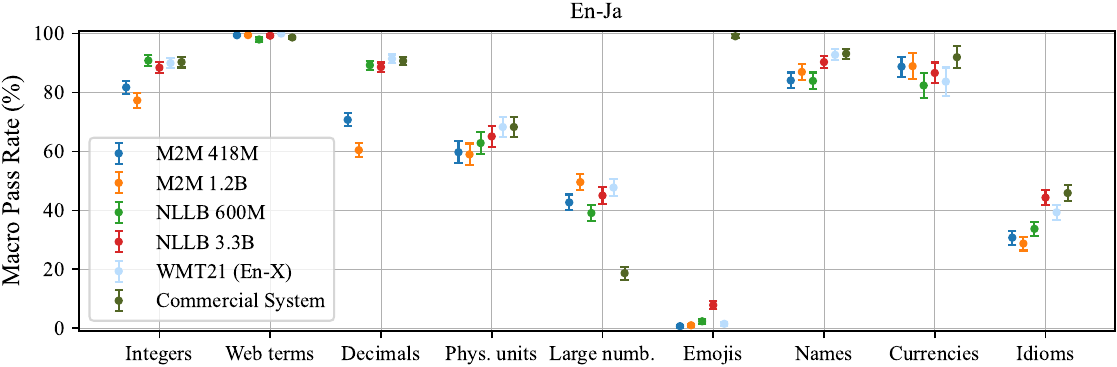}
 \captionof{figure}{Macro pass rate confidence intervals for En$\rightarrow$Ja tests.}
\label{fig:en_ja_MPR.png}
\end{minipage}

\noindent\begin{minipage}{\textwidth}
\vspace{5ex}
\centering
\renewcommand{\arraystretch}{1.5}
\resizebox{\linewidth}{!}{
\begin{tabular}{ccccccc}
\toprule
\multirow{2}{*}{\textbf{Property}}                      & \multicolumn{5}{c}{\textbf{Macro Pass Rate} (\%)}         \\ \cmidrule{2-7} 
               & \textbf{M2M 418M}  & \textbf{M2M 1.2B}  & \textbf{NLLB 600M} & \textbf{NLLB 3.3B} & \textbf{WMT21 (En-X)} & \textbf{Commercial system}            \\ \cmidrule{1-7} 
Web terms      & {[}0.991, 0.998{]} & {[}0.992, 0.998{]} & {[}0.973, 0.987{]} & {[}0.989, 0.997{]} & {[}1.0, 1.0{]}        & {[}0.982, 0.992{]}     \\
Decimals       & {[}0.685, 0.73{]}  & {[}0.58, 0.629{]}  & {[}0.878, 0.908{]} & {[}0.871, 0.902{]} & {[}0.9, 0.929{]}      & {[}0.893, 0.922{]}     \\
Integers       & {[}0.795, 0.84{]}  & {[}0.749, 0.798{]} & {[}0.891, 0.926{]} & {[}0.865, 0.904{]} & {[}0.882, 0.918{]}    & {[}0.885, 0.922{]}     \\
Physical Units & {[}0.56, 0.635{]}  & {[}0.553, 0.627{]} & {[}0.591, 0.666{]} & {[}0.615, 0.687{]} & {[}0.648, 0.718{]}    & {[}0.649, 0.717{]}     \\
Large Numbers  & {[}0.4, 0.454{]}   & {[}0.469, 0.523{]} & {[}0.363, 0.419{]} & {[}0.422, 0.479{]} & {[}0.45, 0.505{]}     & {[}0.165, 0.209{]}     \\
Emojis         & {[}0.002, 0.012{]} & {[}0.005, 0.015{]} & {[}0.015, 0.032{]} & {[}0.064, 0.093{]} & {[}0.008, 0.022{]}    & {[}0.984, 0.998{]}     \\
Names          & {[}0.814, 0.868{]} & {[}0.844, 0.896{]} & {[}0.811, 0.868{]} & {[}0.882, 0.925{]} & {[}0.909, 0.947{]}    & {[}0.915, 0.949{]}     \\
Currencies     & {[}0.854, 0.922{]} & {[}0.845, 0.934{]} & {[}0.782, 0.866{]} & {[}0.831, 0.902{]} & {[}0.789, 0.885{]}    & {[}0.883, 0.957{]}     \\
Idioms         & {[}0.284, 0.331{]} & {[}0.264, 0.311{]} & {[}0.313, 0.362{]} & {[}0.418, 0.469{]} & {[}0.368, 0.419{]}    & {[}0.433, 0.485{]}  \\  \bottomrule
\end{tabular}
}
\captionof{figure}{Macro Pass Rate confidence intervals for En$\rightarrow$Ja tests.}
\label{tab:conf_intervals_en_ja}
\end{minipage}

\end{document}

%% file: math_commands.tex
\usepackage{amsmath,amsfonts,bm}

\def\eqref#1{equation~\ref{#1}}

\def\1{\bm{1}}

\def\vc{{\bm{c}}}

\def\vg{{\bm{g}}}

\def\vx{{\bm{x}}}

\DeclareMathAlphabet{\mathsfit}{\encodingdefault}{\sfdefault}{m}{sl}
\SetMathAlphabet{\mathsfit}{bold}{\encodingdefault}{\sfdefault}{bx}{n}



%% file: macros.tex
\usepackage{nameref}
\usepackage{cleveref}

\crefformat{section}{\S#2#1#3}
\crefformat{subsection}{\S#2#1#3}
\crefformat{subsubsection}{\S#2#1#3}
\crefformat{paragraph}{\P#2#1#3}
\crefformat{subparagraph}{\P#2#1#3}
\crefmultiformat{section}{\S#2#1#3}{ and~\S#2#1#3}{, \S#2#1#3}{, and~\S#2#1#3}
\crefmultiformat{subsection}{\S#2#1#3}{ and~\S#2#1#3}{, \S#2#1#3}{, and~\S#2#1#3}
\crefmultiformat{subsubsection}{\S#2#1#3}{ and~\S#2#1#3}{, \S#2#1#3}{, and~\S#2#1#3}
\crefmultiformat{paragraph}{\P\P#2#1#3}{ and~#2#1#3}{, #2#1#3}{, and~#2#1#3}
\crefmultiformat{subparagraph}{\P\P#2#1#3}{ and~#2#1#3}{, #2#1#3}{, and~#2#1#3}
\crefrangeformat{section}{\mbox{\S\S#3#1#4--#5#2#6}}
\crefrangeformat{subsection}{\mbox{\S\S#3#1#4--#5#2#6}}
\crefrangeformat{subsubsection}{\mbox{\S\S#3#1#4--#5#2#6}}
\crefrangeformat{paragraph}{\mbox{\P\P#3#1#4--#5#2#6}}
\crefrangeformat{subparagraph}{\mbox{\P\P#3#1#4--#5#2#6}}
\crefname{part}{Part}{Parts}
\Crefname{part}{Part}{Parts}
\crefname{chapter}{ch.}{ch.}
\Crefname{chapter}{Ch.}{Ch.}
\crefname{footnote}{fn.}{fn.}
\Crefname{footnote}{Fn.}{Fn.}
\crefname{figure}{figure}{figures}
\crefname{subfigure}{figure}{figures}
\Crefname{subfigure}{Figure}{Figures}
\crefname{appsec}{appendix}{appendices}
\Crefname{appsec}{Appendix}{Appendices}
\crefname{algocf}{algorithm}{algorithms}
\Crefname{algocf}{Algorithm}{Algorithms}

\crefname{ExNo}{example}{examples}
\Crefname{ExNo}{Example}{Examples}
\crefname{SubExNo}{example}{examples}
\Crefname{SubExNo}{Example}{Examples}
\crefname{SubSubExNo}{example}{examples}
\Crefname{SubSubExNo}{Example}{Examples}
\crefformat{ExNo}{(#2#1#3)}
\crefformat{SubExNo}{(#2#1#3)}
\crefformat{SubSubExNo}{(#2#1#3)}
\crefrangeformat{ExNo}{\mbox{(#3#1#4--#5#2#6)}}
\crefrangeformat{SubExNo}{\mbox{(#3#1#4--#5#2#6)}}
\crefrangeformat{SubSubExNo}{\mbox{(#3#1#4--#5#2#6)}}
\crefmultiformat{ExNo}{(#2#1#3}{, #2#1#3)}{, #2#1#3}{, #2#1#3)}
\crefmultiformat{SubExNo}{(#2#1#3}{, #2#1#3)}{, #2#1#3}{, #2#1#3)}
\crefmultiformat{SubSubExNo}{(#2#1#3}{, #2#1#3)}{, #2#1#3}{, #2#1#3)}
\crefrangemultiformat{ExNo}{(#3#1#4--#5#2#6}{, #3#1#4--#5#2#6)}{, #3#1#4--#5#2#6}{, #3#1#4--#5#2#6)}
\crefrangemultiformat{SubExNo}{(#3#1#4--#5#2#6}{, #3#1#4--#5#2#6)}{, #3#1#4--#5#2#6}{, #3#1#4--#5#2#6)}
\crefrangemultiformat{SubSubExNo}{(#3#1#4--#5#2#6}{, #3#1#4--#5#2#6)}{, #3#1#4--#5#2#6}{, #3#1#4--#5#2#6)}

%% file: prompt_src_no_format.tex
{\fontsize{8}{10}
\begin{Verbatim}[commandchars=+!?]
You are an assistant that generates sentences where only appears one B = {property}.
Don't be repetitive, change the topic and B between sentences. Write every B inside [].
B must happen only once in each sentence and can only contain {property}.

Write 3 examples.

- {Source sentence demonstration #1}
- {Source sentence demonstration #2}
- {Source sentence demonstration #3}

Now write 10 more diverse sentences itemizing them with '-':
\end{Verbatim}
\vspace{-0.4cm}
}

%% file: prompt_candidates_no_format.tex
{\fontsize{8}{10}
\begin{Verbatim}[commandchars=+!?]
You are a {source_lang}-{target_lang} translator. Given a {property}, write as many valid {target_lang}
translations as you can. Use "|" to separate between valid translations.
Write "NA" if unable to accomplish the task.

{Source property demonstration #1}
{Candidates set source property demonstration #1}
{Source property #2}
{Candidates set source property demonstration #2}
{Source property #3}
{Candidates set source property demonstration #3}

{Source property}
\end{Verbatim}
\vspace{-0.4cm}
}

%% file: prompt_candidates_no_format_currency.tex
{\footnotesize
\begin{Verbatim}[commandchars=+!?]
You are a {src_lang}-{tgt_lang} translator. Given a {property}, write as many valid
{target_lang} translations as you can. Use "|" to separate between valid translations.
Write "NA" if unable to accomplish the task.

EUR
€|EUR|Euro

GBP
£|GBP|Pfund|Pfund Sterling|britisches Pfund|Pound Sterling

USD
$|USD|Dollar|US Dollar|amerikanischer Dollar|amerikanische Dollar|US-Dollar

{Source property}
\end{Verbatim}
}

%% file: prompt_candidates_contrastive_no_format.tex
Foil:
{\footnotesize
\begin{Verbatim}[commandchars=+!?]
You are an {src_lang}-{tgt_lang} literal translator. Given a sequence of words, 
you have to write only a literal translation. Use "|" to separate alternatives.
Write "NA" if unable to accomplish the task.

break a leg
brich dir ein Bein|breche dir ein Bein|breche dein Bein|breche dir dein Bein

hit the ground running
im Laufen hinfallen|beim Laufen hinfallen|beim Laufen auf den Boden knallen|beim Laufen
auf den Boden fallen

put all your eggs in one basket
alle Eier in einen Korb tun|alle Eier in einen Korb setzen|alle Eier in einen Korb legen

{Source property}
   \end{Verbatim}
}
Correct:
{\footnotesize
\begin{Verbatim}[commandchars=+!?]
You are an {src_lang_name}-{tgt_lang_name} translator of idiomatic expressions. Given an 
idiomatic expression, you have to write the translation of the figurative meaning of the
idiomatic expression. Use "|" to separate alternatives. Write "NA" if unable to accomplish the task.

She told him to “break a leg” just before he went up on stage.
figurative translation of: break a leg
viel Glück|alles Gute|viel Erfolg|du schaffst das|Sie schaffen das

He hit the ground running, so his employer was really happy.
figurative translation of: hit the ground running
voller Begeisterung angehen|enthusiastisch angehen|hart und erfolgreich arbeiten

{Source property}
\end{Verbatim}
}